\documentclass{bmvc2k}
\usepackage{multirow}
\usepackage{float}
\usepackage{amsfonts}
\usepackage{amssymb}
\usepackage{wrapfig}
\usepackage{lipsum}
\usepackage{enumitem}
\title{OpenHuman4D: Open-Vocabulary \\4D Human Parsing}

\addauthor{Keito Suzuki}{k3suzuki@ucsd.edu}{1}
\addauthor{Bang Du}{b7du@ucsd.edu}{1}
\addauthor{Runfa Blark Li}{rul002@ucsd.edu}{1}
\addauthor{Kunyao Chen}{kuc017@ucsd.edu}{2}
\addauthor{Lei Wang}{wlei@qti.qualcomm.com}{2}
\addauthor{Peng Liu}{peli@qti.qualcomm.com}{2}
\addauthor{Ning Bi}{nbi@qti.qualcomm.com}{2}
\addauthor{Truong Nguyen}{tqn001@ucsd.edu}{1}

\addinstitution{
 University of California, San Diego\\
 San Diego, CA, USA
}
\addinstitution{
 Qualcomm\\
 San Diego, CA, USA
}

\runninghead{Student, Prof, Collaborator}{BMVC Author Guidelines}

\begin{document}

\maketitle
\vspace{-4ex}
\begin{abstract}
Understanding dynamic 3D human representation has become increasingly critical in virtual and extended reality applications. However, existing human part segmentation methods are constrained by reliance on closed-set datasets and prolonged inference times, which significantly restrict their applicability. In this paper, we introduce the first 4D human parsing framework that simultaneously addresses these challenges by reducing the inference time and introducing open-vocabulary capabilities. Building upon state-of-the-art open-vocabulary 3D human parsing techniques, our approach extends the support to 4D human-centric video with three key innovations: 1) We adopt mask-based video object tracking to efficiently establish spatial and temporal correspondences, avoiding the necessity of segmenting all frames. 2) A novel Mask Validation module is designed to manage new target identification and mitigate tracking failures. 3) We propose a 4D Mask Fusion module, integrating memory-conditioned attention and logits equalization for robust embedding fusion. 
Extensive experiments demonstrate the effectiveness and flexibility of the proposed method on 4D human-centric parsing tasks, achieving up to 93.3\% acceleration compared to the previous state-of-the-art method, which was limited to parsing fixed classes. 
% Our framework demonstrates superior performance in parsing 4D human videos using an open-vocabulary approach, achieving up to xxx\% acceleration compared to the previous state-of-the-art method, which was limited to parsing fixed classes. Extensive experiments demonstrate the effectiveness and flexibility of the proposed method on 4D human-centric parsing tasks. 
\end{abstract}
\begin{figure*}[t]
    \centering
    % \vspace{-3ex}
    \includegraphics[width=.98\hsize]{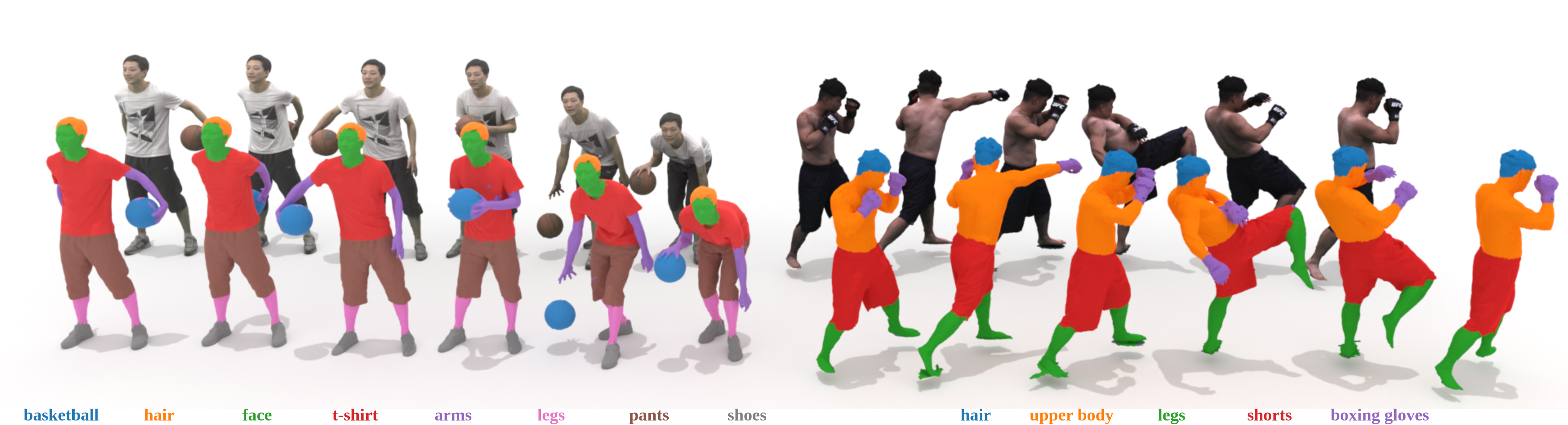}
    \vspace{-1ex}
    \caption{We propose the first open-vocabulary parsing method for 4D human-centric data.} 
    \label{Fig.intro}
    \vspace{-3ex}
\end{figure*}

\vspace{-3ex}
\section{Introduction}
\label{sec:intro}

The development of high-fidelity 3D digital humans is increasingly important across a variety of augmented and virtual reality applications. 3D part segmentation of humans is crucial for understanding a target's structure, surface semantics, mobility, intention, and other key aspects. PartSLIP \cite{liu2023partslip, zhou2023partslip++} leverages 2D priors and the zero-shot capabilities of the image-language model to address the 3D part segmentation task in a zero or few-shot fashion. Find3D \cite{ma2024find} trains a general object part segmentation model to predict point-wise features in a CLIP space for language guided querying.  CloSe-Net \cite{antic2024close} approaches 3D human parsing through labeled dataset and supervised training. \cite{ours3DV} extends the zero-shot capabilities to human-centric data, allowing open vocabulary segmentation on various representations, such as point cloud, mesh, and 3D Gaussians \cite{kerbl20233d}.

While remarkable progress has been made in developing diverse data-driven approaches for 3D part segmentation \cite{antic2024close, ma2024find}, these efforts have primarily focused on static models, lacking support for dynamic sequences, which are the most common representation in human-centric applications. 
Recent work \cite{wang20244d} addresses this gap with 4D clothed human parsing. They propose a parsing pipeline that combines the pre-trained 2D segmentation \cite{kirillov2023segment}, 2D human parsing \cite{gong2019graphonomy}, and optical flow models \cite{teed2020raft} to label 4D human data. 
% By leveraging semantic maps from a 2D human parser \cite{gong2019graphonomy}, they extended these techniques to 4D, accounting for both multi-view and temporal consistency.
Although this approach reduces manual efforts and produces state-of-the-art results, it remains computationally expensive during inference, limiting its applicability to scenarios beyond one-time label generation. Moreover, it only supports fixed classes and cannot effectively handle labels with complex clothing or arbitrary objects interacting with humans. 

To address these issues, we introduce the first open-vocabulary 4D parsing method, enabling efficient and robust handling of arbitrary human-centric data. Our method employs a visual object tracker to directly establish inter-frame and inter-view correspondences for each mask. We complement this with a mask validation module designed to detect tracking failures and handle new targets. This approach eliminates the need for expensive per-frame mask proposals. Moreover, we develop a novel 4D MaskFusion module that fuses features from multiple views and frame at the same time. The module enables current frame feature generation to condition on embeddings from other frames and views. By performing self-attention across memorized embeddings, we enhance the robustness of the parsing process.

We evaluate our method against state-of-the-art 3D and 4D approaches by performing per-frame segmentation on the CTD Dynamic \cite{chen2021tightcap} and MPEG-PCC \cite{xu2017owlii} datasets. Experimental results show that our method outperforms all competing methods in numerical evaluation metrics. Our method produces temporally coherent parsing results with the ability to segment finer parts and objects compared to the current state-of-the-art 4D method. At the same time, our proposed approach delivers a substantial efficiency improvement, accelerating inference speed by up to 93.3\%.

In summary, the key contributions of our approach are: 
\begin{itemize}[topsep=1.5pt, parsep=0.8pt]
    \item We propose the first open-vocabulary 4D human-centric segmentation method achieving more accurate and faster inference than the state-of-the-art 3D and 4D methods.
    \item We introduce an efficient 4D mask proposal generator by adopting a visual object tracker and a novel mask validation module.
    \item We design a novel 4D MaskFusion module with memory of inter-frame correspondences for more robust segmentation.
\end{itemize}

% The development of high-fidelity 3D digital humans is increasingly important across a variety of
% augmented and virtual reality applications. 3D part segmentation on human, which is crucial in understanding target's structure, surface semantics, mobility, intention, etc. Remarkable progress has been made in developing diverse data-driven approaches for 3D part segmentation, but majorly focused on static 3D models, lacking the support of animated sequences, which is the mostly common representation in human-centric applications. 

% recent work of 4d human parsing (4d dress) and our work on 3D Human parts segmentation. extending from it

% previous 4d-dress propose a parsing pipeline combines SAM and optical flow to label the human data. developed a semi-automatic
% and template-free 4D human parsing pipeline. Leveraging semantic maps from a 2D human parser  and a segmentation model , we extended these techniques to 4D,
% considering both multi-view and temporal consistency. 
% Although it largely minimizes the manual efforts and produces the SOTA results, it is costly and hard to be applied in applications beyond label preparation. Also it only supports fixed classes, cannot handle complex clothes and arbitrary object interacted with human
\vspace{-3ex}
\section{Related Works}
\vspace{-1ex}
\label{sec:related_works}

% \begin{figure}[t]
%     \centering
%     \includegraphics[width=.8\hsize]{figures/single_frame_problems.png}
%     \caption{Failure cases of static 3D human parsing \cite{ours3DV}. While some frames (1 and 3) have accurate segmentation, other frames (2 and 4) can have noisy segmentation results.}
%     \vspace{-3ex}
%     \label{Fig.single_frame_negative}
% \end{figure}
\noindent
\textbf{3D Human Paring. }
Early works in static 3D human parsing \cite{jertec2019using, ueshima2021training} primarily focused on parsing body parts by training a point cloud segmentation network \cite{qi2017pointnet} on data created with parametric human models. GIM3D \cite{musoni2022gim3d, musoni2023gim3d} trained 3D networks \cite{qi2017pointnet, qi2017pointnet++, sharp2022diffusionnet} on synthetic data from clothing simulations \cite{bertiche2020cloth3d}. CloSe-Net \cite{antic2024close} achieves SOTA results in 3D clothing segmentation by leveraging a human prior \cite{SMPL:2015} and a garment-based attention module. They also release labels for publicly available 3D human datasets \cite{jinka2023sharp, cai2022humman, yu2021function4d, zheng2022structured}. Most recently, \cite{ours3DV} proposed an open-vocabulary approach through foundation models \cite{ravi2024sam, radford2021learning, sun2024alpha}. Directly extending these approaches to 4D sequences is difficult as they lack temporal information and can be inefficient.

\noindent
\textbf{4D Human Parsing. }
While there have been both synthetic \cite{guan2012drape, pumarola20193dpeople, zou2023cloth4d, black2023bedlam} and real \cite{ClothCap2017, Zhang_2017_CVPR, shen2023x} 4D human datasets, most lacked labeled data to train or evaluate 4D human parsing models. An early work \cite{ClothCap2017} parses scans by registering a SMPL model \cite{SMPL:2015} with per-vertex offsets. The most recent work, 4D-DRESS \cite{wang20244d}, proposes a semi-automatic pipeline to parse human mesh sequences using a multi-view approach. They leverage pretrained 2D models to project 2D labels onto 3D surfaces and can predict temporally consistent segmentation results for 4D humans. Furthermore, they manually edit their segmentation results to release the first real-world 4D human parsing dataset. Although their method can obtain accurate results, it is inefficient and limited to parsing into six categories. In this paper, we propose a more efficient approach with adjustable segmentation categories based on user text inputs.

\noindent
\textbf{Open-Vocabulary 3D Part Segmentation. }
With the popularity of vision language models \cite{jia2021scaling, radford2021learning, zhai2023sigmoid, zheng2024dreamlip}, there has been growing success in both open-vocabulary 2D \cite{ding2022decoupling, li2022languagedriven, ghiasi2022scaling, liang2023open, jiao2024collaborative} and 3D segmentation \cite{zhu2024open, yan2024maskclustering, shi2024language, nguyen2024open3dis, takmaz2023openmask3d}. For 3D part segmentation, recent methods \cite{liu2023partslip, zhou2023partslip++, abdelreheem2023satr, zhong2024meshsegmenter} have adopted pre-trained open-set object detectors \cite{li2022grounded, liu2024grounding} to take a multi-view approach. Find3D \cite{ma2024find} recently proposed a transformer based \cite{wu2024point} model that directly estimates point-wise features which can be queried with text prompts. While these models have shown success in general 3D objects and scenes, they have not seen the same success in dynamic 3D humans. Therefore, we propose a 4D human segmentation pipeline with open-vocabulary capabilities.
\vspace{-3ex}
\section{Methodology}
\vspace{-1ex}
\subsection{Preliminary: Open Vocabulary 3D Human Segmentation}
\label{sec:preliminary}
\cite{ours3DV} introduces a framework for open-vocabulary 3D human segmentation (denoted as OpenHuman3D in this paper). It leverages a 3D mask classification and fusion-based approach to synthesize proposals generated through a pretrained segmentation model \cite{kirillov2023segment}. Given a colored point-based representation, it first renders multiple images from $V$ predefined camera poses. Then, the rendered images go through the  following three modules: 

\noindent\textbf{Multi-view Mask Proposal Generation.} 
SAM \cite{kirillov2023segment} is employed to generate 2D mask proposals. Each rendered image $I_i \in \mathbb{R}^{H\times W \times 3}$, where $i \in [1, V]$ denotes the index of the view, is independently fed into SAM to generate $N_i$ class-agnostic masks. This results in a total of $N=\sum_{i=1}^V N_i$ binary 2D masks at varying granularities.

% \noindent\textbf{HumanCLIP Encoding.}
% HumanCLIP is a human-centric AlphaCLIP \cite{sun2024alpha} model that takes the original rendered image and an additional binary mask to effectively encode the part region. Each 2D mask $m_{i, j}^{2D}$, where $m_{i, j}^{2D}$ is the $j$-th 2D mask generated by SAM from the $i$-th view, with its corresponding image $I_i$ is fed to the image encoder to get the proposal CLIP \cite{radford2021learning} embedding $q_{i, j} \in \mathbb{R}^{D}$. The user-defined text prompt is fed into HumanCLIP text encoder to obtain CLIP text embeddings $\mathbf{W} \in \mathbb{R}^{K \times D}$, where $K$ corresponds to the number of segmentation labels.

\noindent\textbf{3D Mask Fusion.}
The 3D mask proposal $m_{i, j}^{3D}$ is formed by unprojecting its corresponding 2D mask proposal $m_{i, j}^{2D}$ using the camera parameters. 
The final proposals $\mathbf{M} \in \{0, 1\}^{P \times N}$ and their embeddings $\mathbf{Q} \in \mathbb{R}^{N \times D}$ are obtained by concatenating the 3D masks and embeddings from all views. 
The classification logits $\mathbf{P} \in \mathbb{R}^{N \times K}$ are computed by taking the cosine similarity between each mask and text embedding.
% \begin{equation}
%     \label{eq:mask_logits}
%     \mathbf{P}_{n, k} = \frac{\mathbf{Q}_n\cdot \mathbf{W}_k}{||\mathbf{Q}_n||||\mathbf{W}_k||}
% \end{equation}
The resulting 3D segmentation result $\mathbf{Y} \in \mathbb{R}^{P \times K}$ is computed as the weighted sum of 3D masks $\mathbf{M}$ based on the classification logits $\mathbf{P}$: $\mathbf{Y} = \mathbf{M} \times \mathbf{P}$. An additional step sets points to a `no label' class if the maximum logits of the points are lower than a threshold $\tau$.

\begin{figure*}[t]
    \centering
    \includegraphics[width=.98\hsize]{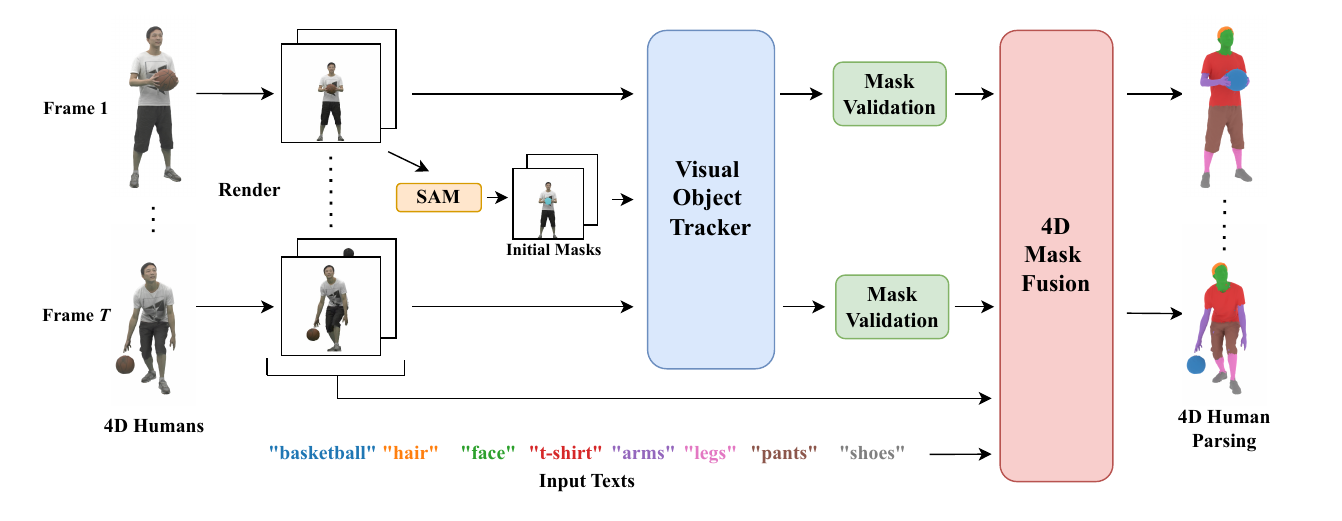}
    \vspace{-3ex}
   \caption{Overview of the proposed framework.} 
    % \caption{Overview of our proposed framework. Given 4D humans, the initial mask proposals are generated with SAM based on the first view of the first frame. These masks along with all rendered images are fed to an object tracker to propagate the masks to all the images. The Mask Validation module then checks and augments the propagated proposals. Finally, the mask proposals along with the multi-view images and input texts are fed to a 4D Mask Fusion module to obtain the final 4D segmentation results.}
    \label{Fig.model}
    \vspace{-3ex}
\end{figure*}

\subsection{Mask Proposal Propagation}
\label{sec:mask_proposal}
Without temporal information, the resulting segmentation often exhibits jittery artifacts and fails in challenging cases. Furthermore, processing each frame independently introduces substantial computational redundancy, as this approach disregards the rich inter-frame information that could otherwise enhance efficiency. We propose to integrate visual object tracking to obtain inter-frame correspondences for smoother and more efficient per-frame mask segmentation. In this work, we utilize SAM 2~\cite{ravi2024sam}, which  demonstrates capabilities in extending SAM's segmentation results by incorporating memory mechanisms for tracking objects across multiple frames in dynamic video sequences. 

Given a sequence of $T$ 3D humans, we first render them from $V$ views to obtain a set of images $\mathcal{I}$ of length $T \times V$:
\begin{equation}
    \mathcal{I} = \{ I_{t,v} \mid t \in \{1,\dots, T\},\; v \in \{1,\dots, V\} \}    
\end{equation}
where $I_{t, v} \in \mathbb{R}^{H\times W \times 3}$ and $t, v$ are the indices for the frame and view respectively.

To generate the initial mask proposals, SAM \cite{kirillov2023segment} is applied to image $I_{1, 1}$ from the first view of the first frame to generate a set of $N$ class-agnostic overlapping masks:
\begin{equation}
    \label{eq:sam}
    [m^1, ..., m^N] = f_{SAM}(I_{1, 1})
\end{equation}
where $m^i \in \{0, 1\}^{H\times W}$ is the $i$-th binary mask generated from SAM.
% where $m^i \in \mathbb{R}^{H\times W}$ is the $i$-th mask generated from SAM.

Given the entire image sequence $\mathcal{I}$ and a mask prompt $m^i$, we apply SAM 2 to propagate this mask to all views and frames to get a set of masks $\mathcal{M}^i$:
\begin{align}
    \mathcal{M}^i &= \{ m_{t,v}^i \mid t \in \{1,\dots, T\},\; v \in \{1,\dots, V\} \} \notag \\
    &= f_{SAM2}(\mathcal{I}, m^i)
\end{align}
where $m_{t, v}^i$ is mask $m^i$ propagated to the image at frame $t$ from view $v$. This is repeated for all of the $N$ masks initially proposed by SAM.

% equationize the input output

% \begin{figure}[t]
%     \centering
%     \includegraphics[width=.7\hsize]{figures/new_masks.png}
%     \vspace{-2ex}
%     % \caption{Failure cases of Visual Object Tracking. In the initial frame, the masks comprehensively cover the entire human subject. In the later frame, coverage gaps become apparent. (a), (b) Initial and current frame with the union of the generated masks overlaid on the image in green. The boxes show areas with no masks. (c) Newly generated masks to compensate for uncovered areas.}
%     \caption{Failure cases of Visual Object Tracking. (a), (b) Initial and current frame with the union of the generated masks overlaid on the image in green. The boxes show areas with no masks. (c) Newly generated masks to compensate for uncovered areas.} 
%     \label{Fig.new_masks}
%     \vspace{-2ex}
% \end{figure}

\subsection{Mask Proposal Validation} 
\label{sec:mask_validation}

\begin{wrapfigure}{r}{0.49\textwidth}
    \centering
    \vspace{-2ex}
    \includegraphics[width=.48\textwidth]{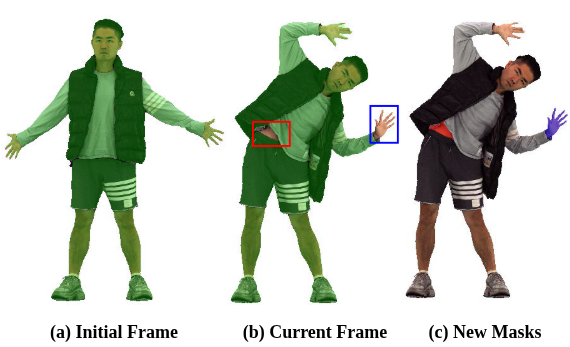}
    \caption{Failure cases of Visual Object Tracking. (a), (b) Initial and current frame with the union of the generated masks overlaid on the image in green. The boxes show areas with no masks. (c) Newly generated masks to compensate for uncovered areas.} 
    \label{Fig.new_masks}
    \vspace{-2ex}
\end{wrapfigure}

Although SAM 2 demonstrates strong performance in basic Visual Object Tracking (VOT), it exhibits two significant limitations: it cannot generate new segmentation masks for regions unseen in the initial frame, and it occasionally fails to propagate existing masks. Figure \ref{Fig.new_masks} illustrates these limitations.

The first limitation is exemplified in the red-boxed region showing the stomach area. Since clothing initially covered this area, no masks from the first frame correspond to the exposed stomach. Consequently, as SAM 2 functions strictly as a tracking model, it lacks the capability to dynamically generate masks for previously unseen regions. Similar issues arise in video sequences when new objects enter the frame. The second limitation is demonstrated in the blue-boxed region, which highlights a hand that was successfully tracked in the initial frame but fails to be detected in the subsequent frame, resulting in lost tracking.

In order to compensate for these incomplete set of masks for each image, we propose a simple mask validation algorithm that detects and adaptively augments uncovered regions with new masks. For an image $I_{t, v}$ at frame $t$, view $v$, we also obtain a binary silhouette mask
$S_{t, v}$ during the rendering process indicating the foreground region. Furthermore, 
to discover the areas covered by the current masks at frame $t$, view $v$, the union of all of the masks is computed $M_{t, v} = \bigcup_{i=1}^N m_{t, v}^i$. Then, the area without any masks $\Tilde{M}_{t, v}$ is determined through the following operation:
\begin{equation}
    \Tilde{M}_{t, v} = S_{t, v} \cap \overline{M}_{t, v}
\end{equation}
As shown in Figure \ref{Fig.new_masks}, the missing area $\Tilde{M}_{t, v}$ may include distinct parts such as the stomach and hand. Therefore, we split $\Tilde{M}_{t, v}$ into $C$ connected components $\mathcal{C}(\Tilde{M}_{t, v})$:
\begin{equation}
    \mathcal{C}(\Tilde{M}_{t, v}) = \{\Tilde{m}_{t, v}^1, ..., \Tilde{m}_{t, v}^C\}
\end{equation}
such that $\Tilde{M}_{t, v} = \bigcup_{i=1}^C m_{t, v}^i$ and $m_{t, v}^i \cap m_{t, v}^j = \varnothing$ for $i \neq j$.
This algorithm results in a new set of masks as shown in red and blue in Figure \ref{Fig.new_masks} (c).

\subsection{4D MaskFusion}
In the MaskFusion module of static 3D human segmentation~\cite{ours3DV}, each mask proposal and its associated embedding are processed independently without any interactions with masks from other views. We observe that due to the lack of inter-view and inter-frame correspondences, it can cause the segmentation quality to greatly vary from frame to frame. We demonstrate the resultant video in the Supplementary. To address this problem, inspired by SAM 2 \cite{ravi2024sam}, we formulate the 4D MaskFusion by integrating a memory bank to explicitly store the inter-frame and inter-view correlation of mask embeddings at the same time. It is accompanied by a memory attention process to simultaneously extract information from all correlated masks across views and frames and generate robust embedding for the current frame. A visualization of the 4D MaskFusion pipeline is shown in the supplementary. 

\noindent\textbf{Fusion: Multi-view Multi-frame All at Once. } For each mask generated in Section \ref{sec:mask_proposal}, we compute the corresponding mask embedding using the HumanCLIP~\cite{ours3DV} vision encoder, creating embedding $q_{i, j} \in \mathbb{R}^{D}$ for $j$-th 2D mask generated by SAM from the $i$-th view. Then, for each $i \in [1, N]$, we construct a memory bank $Q_i \in \mathbb{R}^{(T\times V)\times D}$ where we concatenate all of the mask embeddings from all frames and views originating from the $i$-th mask generated by SAM (Eq. \ref{eq:sam}) and $D$ is the size of the embedding. Since these concatenated embeddings are computed from masks tracked from the same object mask $m_i$, we can reasonably assume that they exhibit relative similarity. Consequently, we update the embeddings through the memory attention based on the pairwise similarity with embeddings from other frames and views. This is represented as:
\begin{equation}
    Q_i' = \text{softmax}(Q_i Q_i^T) Q_i
\end{equation}
where $Q_i'$ is the updated mask embeddings. A benefit of this self-attention mechanism is its inherent ability to assign higher weights to similar masks and lower weights to dissimilar ones. Although SAM 2 is tracking the same object, its accuracy is not guaranteed in all scenarios. A common error is observed in masks covering the limbs as they can flip between left and right parts when propagated to an image from a different view. In such cases, simple averaging would yield ambiguous embeddings with diminished discriminative capacity. Examples are provided in Section \ref{sec:ablation} and the supplementary.

Accordingly, the mask embeddings for those generated in Section \ref{sec:mask_validation} are computed individually, as no correspondences are obtained from a VOT model for these instances. Finally, all of the masks are unprojected to 3D and their corresponding embeddings are gathered for each frame. This results in 3D masks $\mathbf{M}_t \in \{0, 1\}^{P_t \times L_t}$ and embeddings $\mathbf{Q}_t \in \mathbb{R}^{L_t \times D}$ for each frame $t \in [1, T]$ where $P_t$, $L_t$ are the number of points and total number of masks at frame $t$ respectively.

\noindent\textbf{Logits Equalization.} With $\mathbf{M}_t$ and $\mathbf{Q}_t$, we can apply the same computation as shown in Sec. \ref{sec:preliminary} to obtain the segmentation results for each frame. However, we observe that it is susceptible to noisy or ambiguous masks which can be misclassified to an incorrect class even with lower confidence values than other masks classified to the same class. An example of this is illustrated in the supplementary.
% An example of this is shown in Figure \ref{Fig.discrimination}. The figure shows three masks for the shirt, shorts, and a part of the shirt and their corresponding classification logits. For the third mask, it can be seen that it is classified as `shorts' but has lower confidence when compared to the actual shorts mask. This can result in false segmentation results as seen in the top row of Figure \ref{Fig.discrimination}(b). 
To enhance the discrimination ability, we apply a simple trick to further lower the logits for low confidence masks. This is done by computing a min-max normalization of the intra-class logits which are then element wise multiplied to the original logits $P_t \in \mathbb{R}^{L_t \times K}$ to obtain an equalized logits $P_t' \in \mathbb{R}^{L_t \times K}$ where $K$ is the number of classes. The process is expressed as: 
\begin{equation}
\vspace{-1ex}
    \label{eq:logits_eq}
    P_{t, k}' = \frac{P_{t, k} - \min P_{t, k}}{\max P_{t, k} - \min P_{t, k}} \odot P_{t, k}
\end{equation}
where $P_{t, k} \in \mathbb{R}^{L_t}$ is the logits within the $k$-th class.
As a result, we can mitigate the effect of these ambiguous masks to get more accurate segmentation results. 
\vspace{-2ex}
\section{Experiments}
\vspace{-1ex}

\subsection{Implementation Details}
% \noindent\textbf{Models.}
We pick SAM's ViT-H model checkpoint to generate the initial mask proposals. The 
``segment everything" mode is applied to a rendered image with a resolution of $512 \times 512$ where it is prompted with 64 points evenly spaced out along each side of the image. For SAM 2 \cite{ravi2024sam}, we adopt the lightest and the fastest SAM 2.1's ``hiera\_tiny'' model checkpoint. HumanCLIP \cite{ours3DV} is a recently released human-centric vision-language model that finetunes the AlphaCLIP \cite{sun2024alpha} model on 2D human mask-text data.

\begin{table*}[]
    \centering
    \caption{Comparison on the CTD dataset \cite{chen2021tightcap}. OA, mAcc, and mIoU are the overall accuracy, mean class accuracy, and mean Intersection of Union respectively. The best results are shown in \textbf{bold}.}
    \label{tab:quantitative}
    \resizebox{0.8\textwidth}{!}{
    \begin{tabular}{c|ccccc}
    Metric & Find3D \cite{ma2024find} & CloSe-Net \cite{antic2024close} & OpenHuman3D \cite{ours3DV} & 4D-DRESS \cite{wang20244d} & Ours \\ \hline
    OA & 77.97 & 91.48 & 95.68 & 95.65 & \textbf{96.88}\\
    mAcc & 78.09 & 89.01 & 93.20 & 93.08 & \textbf{96.17} \\
    mIoU & 60.59 & 76.40 & 88.96 & 83.55 & \textbf{92.78}
    \end{tabular}
    }
    \vspace{-3ex}
\end{table*}

\subsection{Comparisons}
\noindent\textbf{Dataset.} 
In order to quantitatively evaluate each method, ground truth segmentation labels are required for 4D human data. Recently, 4D-DRESS \cite{wang20244d} released a real-world 4D human dataset, but their annotations were generated using their method with less than 1.5\% manual editing. This makes it difficult to adopt their dataset for evaluation as the ground truth can be a direct output of their method.

Therefore, to evaluate on a more neutral dataset, we generate ground truth labels for the CTD Dynamic dataset \cite{chen2021tightcap}. We annotated the vertices for 10 sequences where each vertex can be classified into 4 categories (`upper clothing', `lower clothing', `shoes', and `other'). The annotation is not done by an existing method and provides a fair comparison for each method. The labels will be publicly released and more details on the generation is provided in the supplementary.

% The dataset consists of sequences of 3D human meshes along with part meshes for the upper clothing, lower clothing, and shoes. By using these part meshes, we automatically label the mesh vertices by assigning it to the class of the part mesh with the closest vertex. If the distance to the closest vertex is larger than a set threshold, it is labeled as an `other' class. This results in a maximum of 4 categories (`upper clothing', `lower clothing', `shoes', and `other') per scan depending on what the person is wearing. We only label the sequences in which the part meshes were provided, resulting in a total of 10 sequences for evaluation. We will release the newly created labels to the public. 

In addition to the CTD Dynamic dataset \cite{chen2021tightcap}, we incorporate the MPEG-PCC dataset \cite{xu2017owlii} for comprehensive visual comparison. Each sequence contains 300 frames, providing an ideal testbed for evaluating performance in scenarios where humans interact with objects. Furthermore, due to the longer video sequences available in MPEG-PCC, we utilize the dataset to conduct comparative analyses of inference times across different methods.

\noindent\textbf{Methods.}
To evaluate the effectiveness of our proposed pipeline, we compare it with four recent segmentation methods: 4D-DRESS \cite{wang20244d}, CloSe-Net \cite{antic2024close}, OpenHuman3D \cite{ours3DV}, and Find3D \cite{ma2024find}.
4D-DRESS \cite{wang20244d} is the state-of-the-art 4D human segmentation method. CloSe-Net \cite{antic2024close} and OpenHuman3D \cite{ours3DV} are the most recent 3D human segmentation methods. Find3D \cite{ma2024find} is a general open-vocabulary part segmentation approach.
% 4D-DRESS \cite{wang20244d} is a state-of-the-art 4D human segmentation method that applies pre-trained 2D human image parsing network \cite{gong2019graphonomy}, optical flow network \cite{teed2020raft}, and SAM \cite{kirillov2023segment} to rendered multi-view images in order to obtain per-pixel candidate labels. These labels are then re-projected to 3D and fused with a Graph Cut optimization to obtain per-vertex segmentation labels. CloSe-Net \cite{antic2024close} is a state-of-the-art 3D human segmentation method trained on a large-scale dataset of 3D human scans. It processes an input point cloud along with corresponding SMPL \cite{SMPL:2015} parameters and clothing category information to segment each point into 18 distinct classes. In addition to these specialized 3D human segmentation methods, we conduct comparisons with open-vocabulary based approaches: OpenHuman3D~\cite{ours3DV}, a recent open-vocabulary 3D human part segmentation method, and Find3D~\cite{ma2024find}, a general part segmentation part segmentation approach. 
Note that except for 4D-DRESS, all compared methods are designed for static objects and are applied independently to the 3D model at each frame. 

\noindent\textbf{Comparison on the CTD Dataset.} 
Table \ref{tab:quantitative} shows the averaged results across 10 sequences from the CTD dataset \cite{chen2021tightcap}. Each method is evaluated using three metrics: overall accuracy (OA), mean class accuracy (mAcc), and mean Intersection of Union (mIoU). Details on each metric as well as the results on individual sequences are included in the supplementary. From the table, it can be observed that the proposed method outperforms the current state-of-the-art 3D and 4D human parsing methods across all three metrics.

\begin{figure*}[t]
    \centering
    \includegraphics[width=.9\hsize]{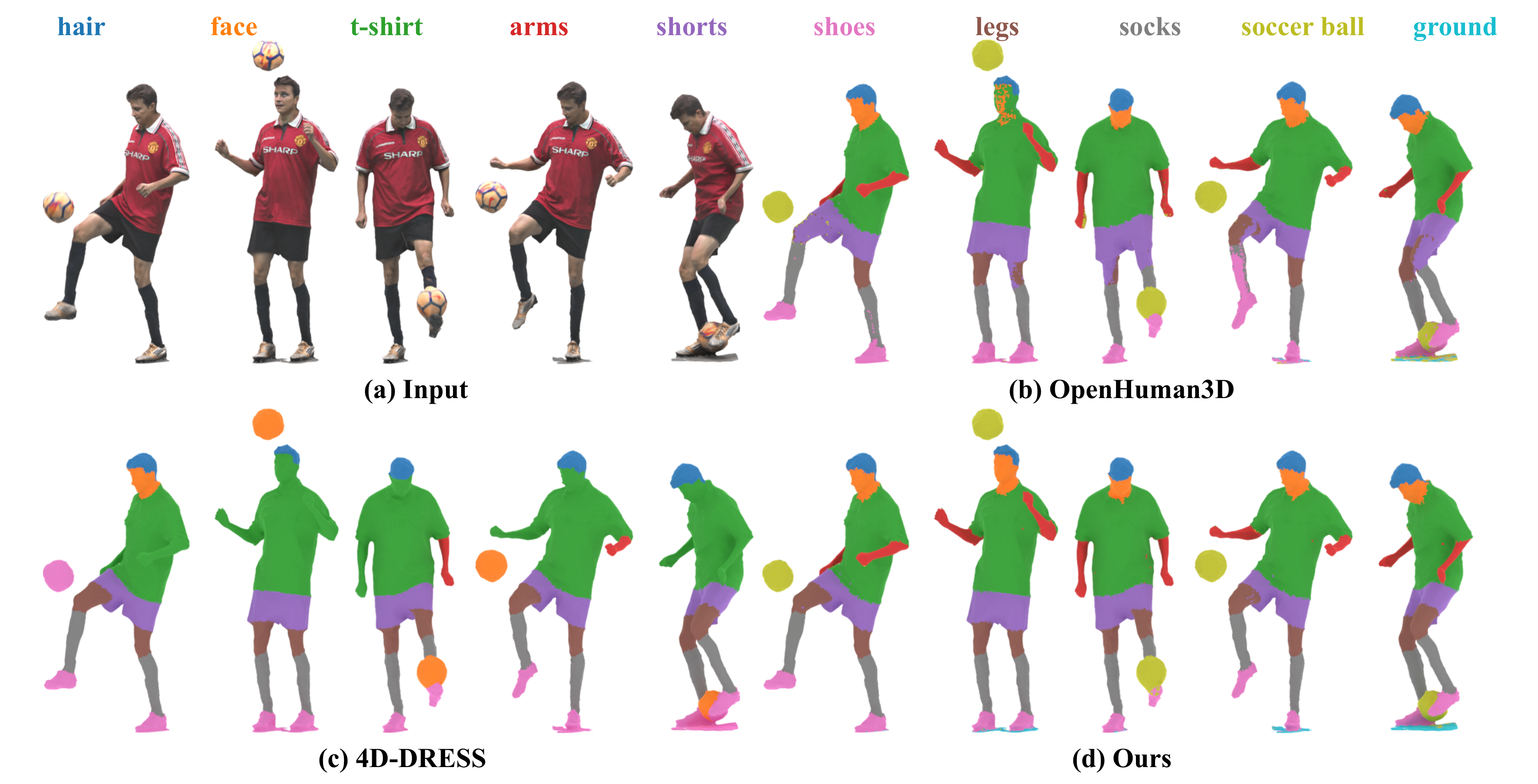}
    \vspace{-2ex}
    \caption{Qualitative comparison on the `football' sequence from the MPEG-PCC dataset \cite{xu2017owlii}. The text at the top shows the corresponding classes and colors.} 
    \vspace{-1ex}
    \label{Fig.football}
\end{figure*}

\noindent\textbf{Visual Comparison on the MPEG-PCC Dataset \cite{xu2017owlii}.} 
Figure \ref{Fig.football} shows a visual comparison with OpenHuman3D and 4D-DRESS on the `football' sequence from the MPEG dataset. 
% We exclude CloSe-Net from the comparison as it requires a registered SMPL model as input, which can be difficult to obtain using off-the-shelf methods when the frame contains objects. 
4D-DRESS adopts a pre-trained Graphonomy model \cite{gong2019graphonomy} which segments an image into 20 categories. The original method merges the body part categories and can only segment into a single `skin' class. To gauge their ability to parse into finer body parts, we relax this class merging process to include face, arms, and legs. 

When comparing the visual results, it is evident that our method produces the most consistent results throughout the entire sequence. Our method successfully segments body parts and clothing, as well as objects and environmental elements such as the soccer ball and ground. OpenHuman3D exhibits inconsistent results across different viewpoints and frames. For 4D-DRESS, it is not expected to segment classes such as `soccer ball' or `ground' as it is not included in the taxonomy of their pre-trained model \cite{gong2019graphonomy}. However, the method shows limited capability in segmenting finer body parts.
% While body parts are accurately parsed in the initial frames, as the sequence progresses, it begins to merge the upper body into a single class and erroneously transfers the `face' class onto the ball. 
The full video sequences are provided in the supplementary.

\noindent\textbf{Inference Time.}
Table \ref{tab:efficiency} shows a comparison of inference time among various methods when segmenting a video sequence with 300 frames. The comparison is conducted on a machine with a single 24 GB RTX 4090 graphics card. The middle three columns of the table shows the input requirements. Find3D is a single frame method and takes a low resolution point cloud as input and achieves the fastest inference time. CloSe-Net is also a single frame method and can directly estimate the segmentation labels for the entire point cloud. However, their model requires preparing a SMPL \cite{SMPL:2015} registration as input, which is not counted in the inference time here. 
% OpenHuman3D applies SAM to every view in every frame and results in the longest inference time. 4D-DRESS takes the entire 3D sequence as input and also applies SAM to rendered images to generate multi-view segmentation results. This process is costly and takes approximately 64\% of the total inference time. 
Both OpenHuman3D and 4D-DRESS apply SAM to every view in every frame which is costly and requires several hours for inferencing a single video sequence.
In contrast, the proposed method integrates a visual object tracker, which efficiently generates mask proposals for each frame, to achieve much faster inference time in comparison to the state-of-the-art 4D human segmentation method.
 
\begin{table}[t]
    \centering
    \caption{Comparison of inference times. We compare inference times for single inference and 100 inference calls. `Direct 3D', `SMPL', and `4D' refer to methods that directly parse in 3D, require SMPL registration, and allow 4D video input respectively.}
    \label{tab:efficiency}
    \resizebox{0.8\columnwidth}{!}{\begin{tabular}{c|ccc|ccc}
         Method & Direct 3D & SMPL & 4D & One-time Inference & Avg. Inference \\ \hline
         Find3D \cite{ma2024find} & \checkmark & & & 7.43s & 0.35s\\
         CloSe-Net \cite{antic2024close} & \checkmark & \checkmark & & 2m 50.00s & 2m 50.00s\\
         OpenHuman3D \cite{ours3DV} & & & & 6h 51m 1.85s & 4m 8.24s\\
         4D-DRESS \cite{wang20244d} & & & \checkmark & 4h 16m 53.99s & 4h 16m 53.99s \\
         Ours & & & \checkmark & 17m 17.90s & 11.75s 
    \end{tabular}
    }
    \vspace{-3ex}
\end{table}

\begin{wrapfigure}{r}{0.49\textwidth}
    \centering
    % \vspace{-4ex}
\includegraphics[width=.48\textwidth]{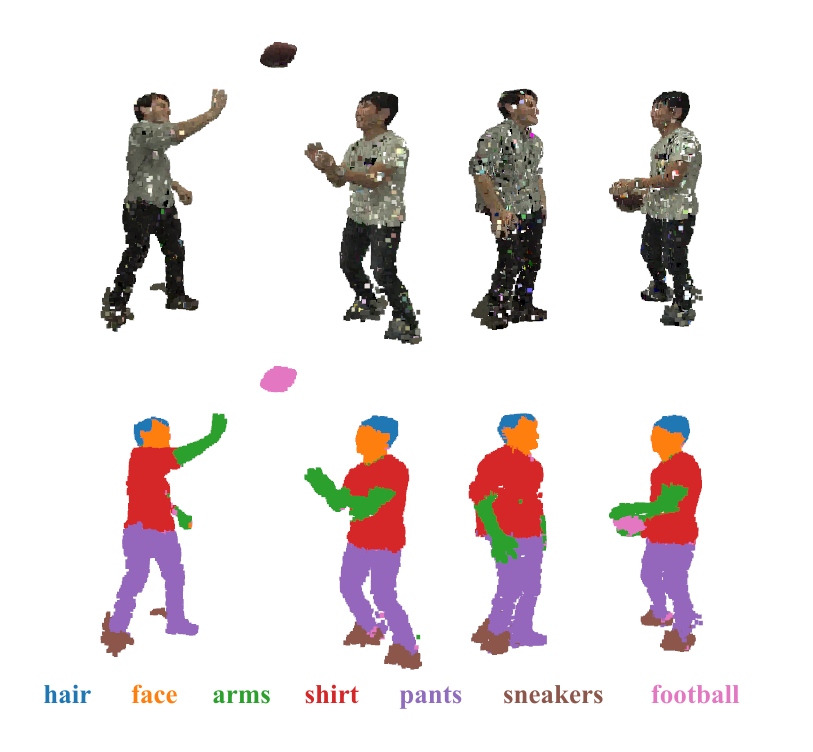}
    \vspace{-3ex}
    \caption{Segmentation result on Dynamic 3D Gaussians \cite{luiten2024dynamic}.} 
    \label{Fig.gaussian_seg}
    \vspace{-3ex}
\end{wrapfigure}

Our design inherits the benefit of OpenHuman3D~\cite{ours3DV} by decoupling user-defined text prompts from mask proposal generation. During the inference stage, only the text encoder and 4D Mask Fusion module remain active, resulting in significantly faster average inference times for the same 4D assets. This efficiency is particularly valuable in VR/XR applications, where users typically input various prompts for identical 4D assets. This advantage is quantitatively demonstrated in the last column of Table~\ref{tab:efficiency}, where our proposed method achieves the second fastest average time across 100 inferences. The results indicate that for the same 4D asset, our method can provide near real-time inference of user-entered prompts while generating more reliable labels compared to Find3D, the fastest. 

\noindent
\textbf{4D Gaussian Segmentation.}
We demonstrate our method's capability in supporting 3D Gaussian sequence, a recently popular way to represent 4D videos. 
Figure \ref{Fig.gaussian_seg} shows an example of applying the proposed framework to parse pretrained Dynamic 3D Gaussian data~\cite{luiten2024dynamic}. The figure exhibits two frames of the `football' sequence from the CMU Panoptic Studio dataset \cite{joo2015panoptic}. The pretrained 3D Gaussians are used to render the images which may not be as high quality as those obtained from curated meshes. However, the results show that our method can still segment these alternative point based representations. Furthermore, it also displays our method's ability to segment multiple people in a single sequence. 
% \begin{figure}[t]
%     \centering
%     \includegraphics[width=.98\hsize]{figures/gaussian_seg2.pdf}
%     \vspace{-2ex}
%     \caption{Segmentation result on Dynamic 3D Gaussians \cite{luiten2024dynamic}.} 
%     \label{Fig.gaussian_seg}
%     \vspace{-2ex}
% \end{figure}
\vspace{-2ex}
\subsection{Ablation Study}
\label{sec:ablation}
\begin{figure}[t]
    \centering
    \begin{minipage}[t]{0.49\textwidth}
        \centering
        \includegraphics[width=\linewidth]{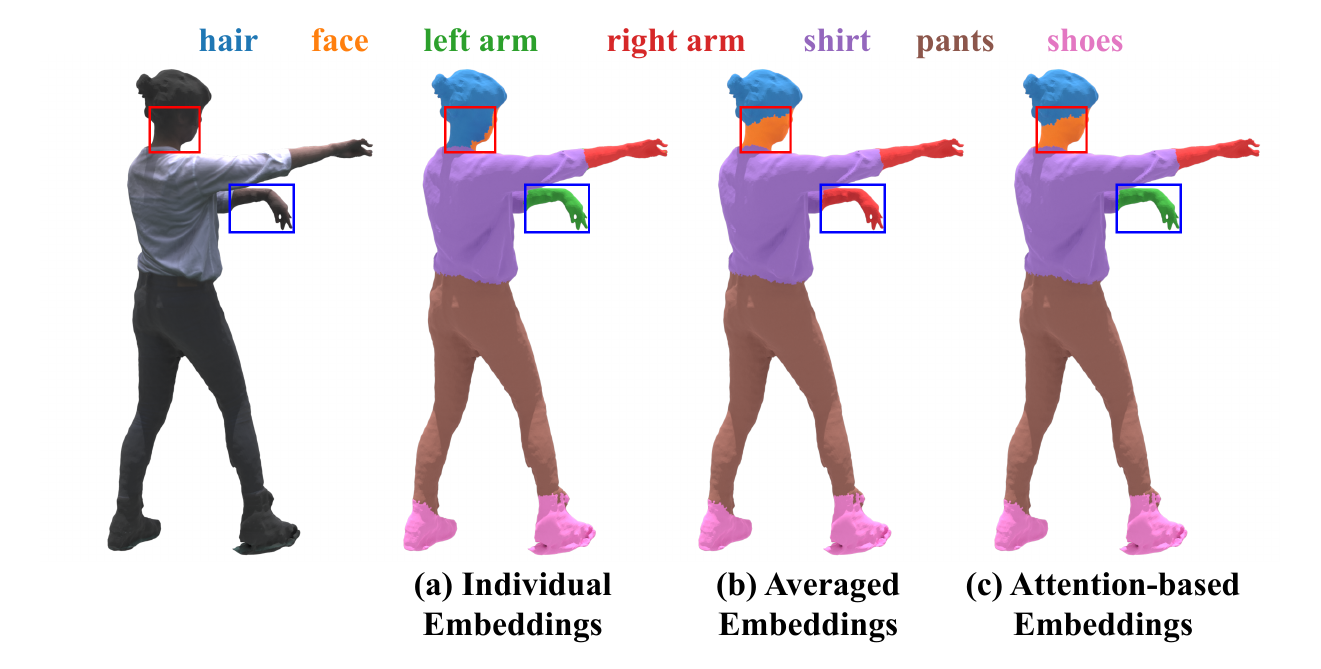}
        \vspace{-4ex}
        \caption{Comparison of segmentation results from different mask embedding computations.}
        \label{Fig.embeddings_ablation}
    \end{minipage}
    % \hfill
    \begin{minipage}[t]{0.49\textwidth}
        \centering
        \includegraphics[width=\linewidth]{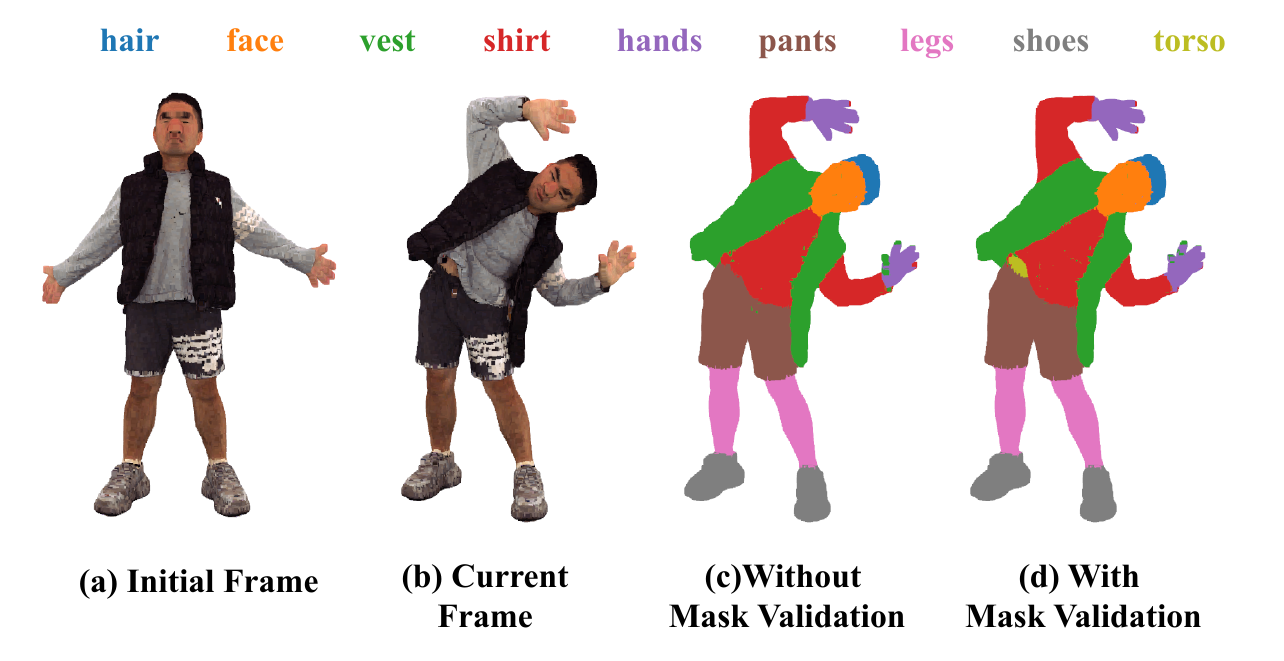}
        \vspace{-4ex}
        \caption{Comparison of segmentation results (c) without mask validation and (d) with mask validation.}
        \label{Fig.mask_validation_ablation}
    \end{minipage}
    \vspace{-2ex}
\end{figure}

\noindent
\textbf{Mask Embedding Computation. }
In the 4D MaskFusion, we integrate embeddings from all views and frames via a memory attention mechanism. We show the effectiveness of this approach in comparison to using individual embeddings without memory as well as simple averaging of embeddings from the memory bank. A comparison of the segmentation result is shown in Figure \ref{Fig.embeddings_ablation}. When using individual embeddings (a), it can result in view inconsistent embeddings, resulting in wrong segmentation of the neck area shown in the red box. When using averaged embeddings (b), it succeeds when the parts are accurately tracked, but can fail when the tracking is inaccurate as shown in the blue box. This is commonly observed in the limbs where masks can flip between left and right parts. In contrast, our attention-based method (c) can effectively integrate embeddings from all views and frames to get a more discriminative representation.

% \begin{figure}[t]
%     \centering
%     % \vspace{-2ex}
%     \includegraphics[width=.8\hsize]{figures/gaussian_seg.pdf}
%     % \vspace{-1ex}
%     \caption{Segmentation result on Dynamic 3D Gaussians \cite{luiten2024dynamic}.} 
%     \label{Fig.gaussian_seg}
% \end{figure}

% \begin{figure}[t]
%     \centering
%     \includegraphics[width=.6\hsize]{figures/embeddings_ablation.pdf}
%     \caption{Comparison of segmentation results from different mask embedding computations.} 
%     \label{Fig.embeddings_ablation}
% \end{figure}

\noindent
\textbf{Effect of Mask Validation. }
The mask validation module serves as a way to introduce novel masks unseen in the initial frame. Figure \ref{Fig.mask_validation_ablation} shows the effectiveness of this module in the final segmentation results. Since the torso region is covered in the initial frame, it is unable to identify this part without the mask validation module. 

% \begin{figure}[t]
%     \centering
%     \includegraphics[width=.6\hsize]{figures/mask_validation_ablation.pdf}
%     \caption{Comparison of segmentation results (c) without mask validation and (d) with mask validation.} 
%     \label{Fig.mask_validation_ablation}
%     \vspace{-3ex}
% \end{figure}

% \noindent
% \textbf{Limitation and Future Works. } While the proposed approach achieves significantly faster inference than the state-of-the-art method, it still does not achieve real-time performance without preprocessing. The future work can be further minimize the reliance on expensive pre-trained models to achieve real-time capabilities for dynamic 4D streams. 

\vspace{-2ex}
\section{Conclusion}
In this paper, we present an efficient solution to 4D human parsing with open-vocabulary capabilities. We adopt a video object tracker for fast mask proposal generation and establishing spatio-temporal correspondences. We introduce a novel Mask Validation module and a 4D MaskFusion module to fuse user-given text inputs for robust segmentation. 
The experimental results show that our method outperforms state-of-the-art 3D and 4D methods in 4D human segmentation.

\newpage

\bibliography{bmvc_final.bbl}

\begin{thebibliography}{55}
\providecommand{\natexlab}[1]{#1}
\providecommand{\url}[1]{\texttt{#1}}
\expandafter\ifx\csname urlstyle\endcsname\relax
  \providecommand{\doi}[1]{doi: #1}\else
  \providecommand{\doi}{doi: \begingroup \urlstyle{rm}\Url}\fi

\bibitem[Abdelreheem et~al.(2023)Abdelreheem, Skorokhodov, Ovsjanikov, and
  Wonka]{abdelreheem2023satr}
Ahmed Abdelreheem, Ivan Skorokhodov, Maks Ovsjanikov, and Peter Wonka.
\newblock Satr: Zero-shot semantic segmentation of 3d shapes.
\newblock In \emph{Proceedings of the IEEE/CVF International Conference on
  Computer Vision}, pages 15166--15179, 2023.

\bibitem[Antić et~al.(2024)Antić, Tiwari, Ozcomlekci, Marin, and
  Pons-Moll]{antic2024close}
Dimitrije Antić, Garvita Tiwari, Batuhan Ozcomlekci, Riccardo Marin, and
  Gerard Pons-Moll.
\newblock {CloSe}: A {3D} clothing segmentation dataset and model.
\newblock In \emph{International Conference on 3D Vision (3DV)}, March 2024.

\bibitem[Bertiche et~al.(2020)Bertiche, Madadi, and
  Escalera]{bertiche2020cloth3d}
Hugo Bertiche, Meysam Madadi, and Sergio Escalera.
\newblock Cloth3d: clothed 3d humans.
\newblock In \emph{European Conference on Computer Vision}, pages 344--359.
  Springer, 2020.

\bibitem[Black et~al.(2023)Black, Patel, Tesch, and Yang]{black2023bedlam}
Michael~J Black, Priyanka Patel, Joachim Tesch, and Jinlong Yang.
\newblock Bedlam: A synthetic dataset of bodies exhibiting detailed lifelike
  animated motion.
\newblock In \emph{Proceedings of the IEEE/CVF Conference on Computer Vision
  and Pattern Recognition}, pages 8726--8737, 2023.

\bibitem[Cai et~al.(2022)Cai, Ren, Zeng, Lin, Yu, Wang, Fan, Gao, Yu, Pan,
  et~al.]{cai2022humman}
Zhongang Cai, Daxuan Ren, Ailing Zeng, Zhengyu Lin, Tao Yu, Wenjia Wang,
  Xiangyu Fan, Yang Gao, Yifan Yu, Liang Pan, et~al.
\newblock Humman: Multi-modal 4d human dataset for versatile sensing and
  modeling.
\newblock In \emph{European Conference on Computer Vision}, pages 557--577.
  Springer, 2022.

\bibitem[Chen et~al.(2021)Chen, Pang, Yang, Wang, Xu, and Yu]{chen2021tightcap}
Xin Chen, Anqi Pang, Wei Yang, Peihao Wang, Lan Xu, and Jingyi Yu.
\newblock Tightcap: 3d human shape capture with clothing tightness field.
\newblock \emph{ACM Transactions on Graphics (TOG)}, 41\penalty0 (1):\penalty0
  1--17, 2021.

\bibitem[Ding et~al.(2022)Ding, Xue, Xia, and Dai]{ding2022decoupling}
Jian Ding, Nan Xue, Gui-Song Xia, and Dengxin Dai.
\newblock Decoupling zero-shot semantic segmentation.
\newblock In \emph{Proceedings of the IEEE/CVF Conference on Computer Vision
  and Pattern Recognition}, pages 11583--11592, 2022.

\bibitem[Ghiasi et~al.(2022)Ghiasi, Gu, Cui, and Lin]{ghiasi2022scaling}
Golnaz Ghiasi, Xiuye Gu, Yin Cui, and Tsung-Yi Lin.
\newblock Scaling open-vocabulary image segmentation with image-level labels.
\newblock In \emph{European Conference on Computer Vision}, pages 540--557.
  Springer, 2022.

\bibitem[Gong et~al.(2019)Gong, Gao, Liang, Shen, Wang, and
  Lin]{gong2019graphonomy}
Ke~Gong, Yiming Gao, Xiaodan Liang, Xiaohui Shen, Meng Wang, and Liang Lin.
\newblock Graphonomy: Universal human parsing via graph transfer learning.
\newblock In \emph{Proceedings of the IEEE/CVF conference on computer vision
  and pattern recognition}, pages 7450--7459, 2019.

\bibitem[Guan et~al.(2012)Guan, Reiss, Hirshberg, Weiss, and
  Black]{guan2012drape}
Peng Guan, Loretta Reiss, David~A Hirshberg, Alexander Weiss, and Michael~J
  Black.
\newblock Drape: Dressing any person.
\newblock \emph{ACM Transactions on Graphics (ToG)}, 31\penalty0 (4):\penalty0
  1--10, 2012.

\bibitem[Jertec et~al.(2019)Jertec, Bojani{\'c}, Bartol, Pribani{\'c},
  Petkovi{\'c}, and Petrak]{jertec2019using}
Andrej Jertec, David Bojani{\'c}, Kristijan Bartol, Tomislav Pribani{\'c},
  Tomislav Petkovi{\'c}, and Slavenka Petrak.
\newblock On using pointnet architecture for human body segmentation.
\newblock In \emph{2019 11th International Symposium on Image and Signal
  Processing and Analysis (ISPA)}, pages 253--257. IEEE, 2019.

\bibitem[Jia et~al.(2021)Jia, Yang, Xia, Chen, Parekh, Pham, Le, Sung, Li, and
  Duerig]{jia2021scaling}
Chao Jia, Yinfei Yang, Ye~Xia, Yi-Ting Chen, Zarana Parekh, Hieu Pham, Quoc Le,
  Yun-Hsuan Sung, Zhen Li, and Tom Duerig.
\newblock Scaling up visual and vision-language representation learning with
  noisy text supervision.
\newblock In \emph{International conference on machine learning}, pages
  4904--4916. PMLR, 2021.

\bibitem[Jiao et~al.(2024)Jiao, Zhu, Huang, Zhao, Wei, and
  Shi]{jiao2024collaborative}
Siyu Jiao, Hongguang Zhu, Jiannan Huang, Yao Zhao, Yunchao Wei, and Humphrey
  Shi.
\newblock Collaborative vision-text representation optimizing for
  open-vocabulary segmentation.
\newblock In \emph{European Conference on Computer Vision}, pages 399--416.
  Springer, 2024.

\bibitem[Jinka et~al.(2023)Jinka, Srivastava, Pokhariya, Sharma, and
  Narayanan]{jinka2023sharp}
Sai~Sagar Jinka, Astitva Srivastava, Chandradeep Pokhariya, Avinash Sharma, and
  PJ~Narayanan.
\newblock Sharp: Shape-aware reconstruction of people in loose clothing.
\newblock \emph{International Journal of Computer Vision}, 131\penalty0
  (4):\penalty0 918--937, 2023.

\bibitem[Joo et~al.(2015)Joo, Liu, Tan, Gui, Nabbe, Matthews, Kanade, Nobuhara,
  and Sheikh]{joo2015panoptic}
Hanbyul Joo, Hao Liu, Lei Tan, Lin Gui, Bart Nabbe, Iain Matthews, Takeo
  Kanade, Shohei Nobuhara, and Yaser Sheikh.
\newblock Panoptic studio: A massively multiview system for social motion
  capture.
\newblock In \emph{Proceedings of the IEEE international conference on computer
  vision}, pages 3334--3342, 2015.

\bibitem[Kerbl et~al.(2023)Kerbl, Kopanas, Leimk{\"u}hler, and
  Drettakis]{kerbl20233d}
Bernhard Kerbl, Georgios Kopanas, Thomas Leimk{\"u}hler, and George Drettakis.
\newblock 3d gaussian splatting for real-time radiance field rendering.
\newblock \emph{ACM Trans. Graph.}, 42\penalty0 (4):\penalty0 139--1, 2023.

\bibitem[Kirillov et~al.(2023)Kirillov, Mintun, Ravi, Mao, Rolland, Gustafson,
  Xiao, Whitehead, Berg, Lo, et~al.]{kirillov2023segment}
Alexander Kirillov, Eric Mintun, Nikhila Ravi, Hanzi Mao, Chloe Rolland, Laura
  Gustafson, Tete Xiao, Spencer Whitehead, Alexander~C Berg, Wan-Yen Lo, et~al.
\newblock Segment anything.
\newblock In \emph{Proceedings of the IEEE/CVF international conference on
  computer vision}, pages 4015--4026, 2023.

\bibitem[Li et~al.(2022{\natexlab{a}})Li, Weinberger, Belongie, Koltun, and
  Ranftl]{li2022languagedriven}
Boyi Li, Kilian~Q Weinberger, Serge Belongie, Vladlen Koltun, and Rene Ranftl.
\newblock Language-driven semantic segmentation.
\newblock In \emph{International Conference on Learning Representations},
  2022{\natexlab{a}}.
\newblock URL \url{https://openreview.net/forum?id=RriDjddCLN}.

\bibitem[Li et~al.(2022{\natexlab{b}})Li, Zhang, Zhang, Yang, Li, Zhong, Wang,
  Yuan, Zhang, Hwang, et~al.]{li2022grounded}
Liunian~Harold Li, Pengchuan Zhang, Haotian Zhang, Jianwei Yang, Chunyuan Li,
  Yiwu Zhong, Lijuan Wang, Lu~Yuan, Lei Zhang, Jenq-Neng Hwang, et~al.
\newblock Grounded language-image pre-training.
\newblock In \emph{Proceedings of the IEEE/CVF conference on computer vision
  and pattern recognition}, pages 10965--10975, 2022{\natexlab{b}}.

\bibitem[Liang et~al.(2023)Liang, Wu, Dai, Li, Zhao, Zhang, Zhang, Vajda, and
  Marculescu]{liang2023open}
Feng Liang, Bichen Wu, Xiaoliang Dai, Kunpeng Li, Yinan Zhao, Hang Zhang,
  Peizhao Zhang, Peter Vajda, and Diana Marculescu.
\newblock Open-vocabulary semantic segmentation with mask-adapted clip.
\newblock In \emph{Proceedings of the IEEE/CVF Conference on Computer Vision
  and Pattern Recognition}, pages 7061--7070, 2023.

\bibitem[Liu et~al.(2023)Liu, Zhu, Cai, Han, Ling, Porikli, and
  Su]{liu2023partslip}
Minghua Liu, Yinhao Zhu, Hong Cai, Shizhong Han, Zhan Ling, Fatih Porikli, and
  Hao Su.
\newblock Partslip: Low-shot part segmentation for 3d point clouds via
  pretrained image-language models.
\newblock In \emph{Proceedings of the IEEE/CVF Conference on Computer Vision
  and Pattern Recognition}, pages 21736--21746, 2023.

\bibitem[Liu et~al.(2024)Liu, Zeng, Ren, Li, Zhang, Yang, Jiang, Li, Yang, Su,
  et~al.]{liu2024grounding}
Shilong Liu, Zhaoyang Zeng, Tianhe Ren, Feng Li, Hao Zhang, Jie Yang, Qing
  Jiang, Chunyuan Li, Jianwei Yang, Hang Su, et~al.
\newblock Grounding dino: Marrying dino with grounded pre-training for open-set
  object detection.
\newblock In \emph{European Conference on Computer Vision}, pages 38--55.
  Springer, 2024.

\bibitem[Loper et~al.(2015)Loper, Mahmood, Romero, Pons-Moll, and
  Black]{SMPL:2015}
Matthew Loper, Naureen Mahmood, Javier Romero, Gerard Pons-Moll, and Michael~J.
  Black.
\newblock {SMPL}: A skinned multi-person linear model.
\newblock \emph{ACM Trans. Graphics (Proc. SIGGRAPH Asia)}, 34\penalty0
  (6):\penalty0 248:1--248:16, October 2015.

\bibitem[Luiten et~al.(2024)Luiten, Kopanas, Leibe, and
  Ramanan]{luiten2024dynamic}
Jonathon Luiten, Georgios Kopanas, Bastian Leibe, and Deva Ramanan.
\newblock Dynamic 3d gaussians: Tracking by persistent dynamic view synthesis.
\newblock In \emph{2024 International Conference on 3D Vision (3DV)}, pages
  800--809. IEEE, 2024.

\bibitem[Ma et~al.(2024)Ma, Yue, and Gkioxari]{ma2024find}
Ziqi Ma, Yisong Yue, and Georgia Gkioxari.
\newblock Find any part in 3d.
\newblock \emph{arXiv preprint arXiv:2411.13550}, 2024.

\bibitem[Musoni et~al.(2022)Musoni, Melzi, Castellani, et~al.]{musoni2022gim3d}
Pietro Musoni, Simone Melzi, Umberto Castellani, et~al.
\newblock Gim3d: A 3d dataset for garment segmentation.
\newblock In \emph{Smart Tools and Applications in Graphics, STAG 2022}, pages
  21--28. 2022.

\bibitem[Musoni et~al.(2023)Musoni, Melzi, and Castellani]{musoni2023gim3d}
Pietro Musoni, Simone Melzi, and Umberto Castellani.
\newblock Gim3d plus: A labeled 3d dataset to design data-driven solutions for
  dressed humans.
\newblock \emph{Graphical Models}, 129:\penalty0 101187, 2023.

\bibitem[Nguyen et~al.(2024)Nguyen, Ngo, Kalogerakis, Gan, Tran, Pham, and
  Nguyen]{nguyen2024open3dis}
Phuc Nguyen, Tuan~Duc Ngo, Evangelos Kalogerakis, Chuang Gan, Anh Tran, Cuong
  Pham, and Khoi Nguyen.
\newblock Open3dis: Open-vocabulary 3d instance segmentation with 2d mask
  guidance.
\newblock In \emph{Proceedings of the IEEE/CVF Conference on Computer Vision
  and Pattern Recognition}, pages 4018--4028, 2024.

\bibitem[Pons-Moll et~al.(2017)Pons-Moll, Pujades, Hu, and Black]{ClothCap2017}
Gerard Pons-Moll, Sergi Pujades, Sonny Hu, and Michael Black.
\newblock Clothcap: Seamless 4d clothing capture and retargeting.
\newblock \emph{ACM Transactions on Graphics, (Proc. SIGGRAPH)}, 36\penalty0
  (4), 2017.
\newblock URL \url{http://dx.doi.org/10.1145/3072959.3073711}.
\newblock Two first authors contributed equally.

\bibitem[Pumarola et~al.(2019)Pumarola, Sanchez-Riera, Choi, Sanfeliu, and
  Moreno-Noguer]{pumarola20193dpeople}
Albert Pumarola, Jordi Sanchez-Riera, Gary Choi, Alberto Sanfeliu, and Francesc
  Moreno-Noguer.
\newblock 3dpeople: Modeling the geometry of dressed humans.
\newblock In \emph{Proceedings of the IEEE/CVF international conference on
  computer vision}, pages 2242--2251, 2019.

\bibitem[Qi et~al.(2017{\natexlab{a}})Qi, Su, Mo, and Guibas]{qi2017pointnet}
Charles~R Qi, Hao Su, Kaichun Mo, and Leonidas~J Guibas.
\newblock Pointnet: Deep learning on point sets for 3d classification and
  segmentation.
\newblock In \emph{Proceedings of the IEEE conference on computer vision and
  pattern recognition}, pages 652--660, 2017{\natexlab{a}}.

\bibitem[Qi et~al.(2017{\natexlab{b}})Qi, Yi, Su, and Guibas]{qi2017pointnet++}
Charles~Ruizhongtai Qi, Li~Yi, Hao Su, and Leonidas~J Guibas.
\newblock Pointnet++: Deep hierarchical feature learning on point sets in a
  metric space.
\newblock \emph{Advances in neural information processing systems}, 30,
  2017{\natexlab{b}}.

\bibitem[Radford et~al.(2021)Radford, Kim, Hallacy, Ramesh, Goh, Agarwal,
  Sastry, Askell, Mishkin, Clark, et~al.]{radford2021learning}
Alec Radford, Jong~Wook Kim, Chris Hallacy, Aditya Ramesh, Gabriel Goh,
  Sandhini Agarwal, Girish Sastry, Amanda Askell, Pamela Mishkin, Jack Clark,
  et~al.
\newblock Learning transferable visual models from natural language
  supervision.
\newblock In \emph{International conference on machine learning}, pages
  8748--8763. PmLR, 2021.

\bibitem[Ravi et~al.(2024)Ravi, Gabeur, Hu, Hu, Ryali, Ma, Khedr, R{\"a}dle,
  Rolland, Gustafson, et~al.]{ravi2024sam}
Nikhila Ravi, Valentin Gabeur, Yuan-Ting Hu, Ronghang Hu, Chaitanya Ryali,
  Tengyu Ma, Haitham Khedr, Roman R{\"a}dle, Chloe Rolland, Laura Gustafson,
  et~al.
\newblock Sam 2: Segment anything in images and videos.
\newblock \emph{arXiv preprint arXiv:2408.00714}, 2024.

\bibitem[Sharp et~al.(2022)Sharp, Attaiki, Crane, and
  Ovsjanikov]{sharp2022diffusionnet}
Nicholas Sharp, Souhaib Attaiki, Keenan Crane, and Maks Ovsjanikov.
\newblock Diffusionnet: Discretization agnostic learning on surfaces.
\newblock \emph{ACM Transactions on Graphics (TOG)}, 41\penalty0 (3):\penalty0
  1--16, 2022.

\bibitem[Shen et~al.(2023)Shen, Guo, Kaufmann, Zarate, Valentin, Song, and
  Hilliges]{shen2023x}
Kaiyue Shen, Chen Guo, Manuel Kaufmann, Juan~Jose Zarate, Julien Valentin, Jie
  Song, and Otmar Hilliges.
\newblock X-avatar: Expressive human avatars.
\newblock In \emph{Proceedings of the IEEE/CVF Conference on Computer Vision
  and Pattern Recognition}, pages 16911--16921, 2023.

\bibitem[Shi et~al.(2024)Shi, Wang, Duan, and Guan]{shi2024language}
Jin-Chuan Shi, Miao Wang, Hao-Bin Duan, and Shao-Hua Guan.
\newblock Language embedded 3d gaussians for open-vocabulary scene
  understanding.
\newblock In \emph{Proceedings of the IEEE/CVF Conference on Computer Vision
  and Pattern Recognition}, pages 5333--5343, 2024.

\bibitem[Sun et~al.(2024)Sun, Fang, Wu, Zhang, Zang, Kong, Xiong, Lin, and
  Wang]{sun2024alpha}
Zeyi Sun, Ye~Fang, Tong Wu, Pan Zhang, Yuhang Zang, Shu Kong, Yuanjun Xiong,
  Dahua Lin, and Jiaqi Wang.
\newblock Alpha-clip: A clip model focusing on wherever you want.
\newblock In \emph{Proceedings of the IEEE/CVF conference on computer vision
  and pattern recognition}, pages 13019--13029, 2024.

\bibitem[Suzuki et~al.(2025)Suzuki, Du, Krishnan, Li, Chen, and
  Nguyen]{ours3DV}
Keito Suzuki, Bang Du, Girish Krishnan, Runfa~Blark Li, Kunyao Chen, and Truong
  Nguyen.
\newblock Open-vocabulary semantic part segmentation of 3d human.
\newblock \emph{arXiv preprint arXiv:2502.19782}, 2025.

\bibitem[Takmaz et~al.(2023)Takmaz, Fedele, Sumner, Pollefeys, Tombari, and
  Engelmann]{takmaz2023openmask3d}
Ay{\c{c}}a Takmaz, Elisabetta Fedele, Robert~W Sumner, Marc Pollefeys, Federico
  Tombari, and Francis Engelmann.
\newblock Openmask3d: Open-vocabulary 3d instance segmentation.
\newblock \emph{arXiv preprint arXiv:2306.13631}, 2023.

\bibitem[Teed and Deng(2020)]{teed2020raft}
Zachary Teed and Jia Deng.
\newblock Raft: Recurrent all-pairs field transforms for optical flow.
\newblock In \emph{Computer Vision--ECCV 2020: 16th European Conference,
  Glasgow, UK, August 23--28, 2020, Proceedings, Part II 16}, pages 402--419.
  Springer, 2020.

\bibitem[Ueshima et~al.(2021)Ueshima, Hotta, Tokai, and
  Zhang]{ueshima2021training}
Takuma Ueshima, Katsuya Hotta, Shogo Tokai, and Chao Zhang.
\newblock Training pointnet for human point cloud segmentation with 3d meshes.
\newblock In \emph{Fifteenth International Conference on Quality Control by
  Artificial Vision}, volume 11794, pages 72--77. SPIE, 2021.

\bibitem[Wang et~al.(2024)Wang, Ho, Guo, Rong, Grigorev, Song, Zarate, and
  Hilliges]{wang20244d}
Wenbo Wang, Hsuan-I Ho, Chen Guo, Boxiang Rong, Artur Grigorev, Jie Song,
  Juan~Jose Zarate, and Otmar Hilliges.
\newblock 4d-dress: A 4d dataset of real-world human clothing with semantic
  annotations.
\newblock In \emph{Proceedings of the IEEE/CVF Conference on Computer Vision
  and Pattern Recognition}, pages 550--560, 2024.

\bibitem[Wu et~al.(2024)Wu, Jiang, Wang, Liu, Liu, Qiao, Ouyang, He, and
  Zhao]{wu2024point}
Xiaoyang Wu, Li~Jiang, Peng-Shuai Wang, Zhijian Liu, Xihui Liu, Yu~Qiao, Wanli
  Ouyang, Tong He, and Hengshuang Zhao.
\newblock Point transformer v3: Simpler faster stronger.
\newblock In \emph{Proceedings of the IEEE/CVF Conference on Computer Vision
  and Pattern Recognition}, pages 4840--4851, 2024.

\bibitem[Xu et~al.(2017)Xu, Lu, and Wen]{xu2017owlii}
Yi~Xu, Yao Lu, and Ziyu Wen.
\newblock Owlii dynamic human mesh sequence dataset.
\newblock ISO/IEC JTC1/SC29/WG11 m41658, 120th MPEG Meeting, Macau, October
  2017.

\bibitem[Yan et~al.(2024)Yan, Zhang, Zhu, and Wang]{yan2024maskclustering}
Mi~Yan, Jiazhao Zhang, Yan Zhu, and He~Wang.
\newblock Maskclustering: View consensus based mask graph clustering for
  open-vocabulary 3d instance segmentation.
\newblock In \emph{Proceedings of the IEEE/CVF Conference on Computer Vision
  and Pattern Recognition}, pages 28274--28284, 2024.

\bibitem[Yu et~al.(2021)Yu, Zheng, Guo, Liu, Dai, and Liu]{yu2021function4d}
Tao Yu, Zerong Zheng, Kaiwen Guo, Pengpeng Liu, Qionghai Dai, and Yebin Liu.
\newblock Function4d: Real-time human volumetric capture from very sparse
  consumer rgbd sensors.
\newblock In \emph{Proceedings of the IEEE/CVF conference on computer vision
  and pattern recognition}, pages 5746--5756, 2021.

\bibitem[Zhai et~al.(2023)Zhai, Mustafa, Kolesnikov, and
  Beyer]{zhai2023sigmoid}
Xiaohua Zhai, Basil Mustafa, Alexander Kolesnikov, and Lucas Beyer.
\newblock Sigmoid loss for language image pre-training.
\newblock In \emph{Proceedings of the IEEE/CVF international conference on
  computer vision}, pages 11975--11986, 2023.

\bibitem[Zhang et~al.(2017)Zhang, Pujades, Black, and
  Pons-Moll]{Zhang_2017_CVPR}
Chao Zhang, Sergi Pujades, Michael~J. Black, and Gerard Pons-Moll.
\newblock Detailed, accurate, human shape estimation from clothed 3d scan
  sequences.
\newblock In \emph{The IEEE Conference on Computer Vision and Pattern
  Recognition (CVPR)}, July 2017.

\bibitem[Zheng et~al.(2024)Zheng, Zhang, Wu, Lu, Ma, Jin, Chen, and
  Shen]{zheng2024dreamlip}
Kecheng Zheng, Yifei Zhang, Wei Wu, Fan Lu, Shuailei Ma, Xin Jin, Wei Chen, and
  Yujun Shen.
\newblock Dreamlip: Language-image pre-training with long captions.
\newblock In \emph{European Conference on Computer Vision}, pages 73--90.
  Springer, 2024.

\bibitem[Zheng et~al.(2022)Zheng, Huang, Yu, Zhang, Guo, and
  Liu]{zheng2022structured}
Zerong Zheng, Han Huang, Tao Yu, Hongwen Zhang, Yandong Guo, and Yebin Liu.
\newblock Structured local radiance fields for human avatar modeling.
\newblock In \emph{Proceedings of the IEEE/CVF Conference on Computer Vision
  and Pattern Recognition}, pages 15893--15903, 2022.

\bibitem[Zhong et~al.(2024)Zhong, Xu, Li, Xu, Li, Yu, and
  Gao]{zhong2024meshsegmenter}
Ziming Zhong, Yanyu Xu, Jing Li, Jiale Xu, Zhengxin Li, Chaohui Yu, and
  Shenghua Gao.
\newblock Meshsegmenter: Zero-shot mesh semantic segmentation via texture
  synthesis.
\newblock In \emph{European Conference on Computer Vision}, pages 182--199.
  Springer, 2024.

\bibitem[Zhou et~al.(2023)Zhou, Gu, Li, Liu, Fang, and Su]{zhou2023partslip++}
Yuchen Zhou, Jiayuan Gu, Xuanlin Li, Minghua Liu, Yunhao Fang, and Hao Su.
\newblock Partslip++: Enhancing low-shot 3d part segmentation via multi-view
  instance segmentation and maximum likelihood estimation.
\newblock \emph{arXiv preprint arXiv:2312.03015}, 2023.

\bibitem[Zhu et~al.(2024)Zhu, Zhou, Xing, Zhao, Xu, Liang, Hauptmann, Liu, and
  Gallagher]{zhu2024open}
Xiaoyu Zhu, Hao Zhou, Pengfei Xing, Long Zhao, Hao Xu, Junwei Liang, Alexander
  Hauptmann, Ting Liu, and Andrew Gallagher.
\newblock Open-vocabulary 3d semantic segmentation with text-to-image diffusion
  models.
\newblock In \emph{European Conference on Computer Vision}, pages 357--375.
  Springer, 2024.

\bibitem[Zou et~al.(2023)Zou, Han, and Wong]{zou2023cloth4d}
Xingxing Zou, Xintong Han, and Waikeung Wong.
\newblock Cloth4d: A dataset for clothed human reconstruction.
\newblock In \emph{Proceedings of the IEEE/CVF Conference on Computer Vision
  and Pattern Recognition}, pages 12847--12857, 2023.

\end{thebibliography}
\end{document}